\documentclass[conference]{IEEEtran}
\IEEEoverridecommandlockouts
\usepackage{cite}
\usepackage{amsmath,amssymb,amsfonts}
\usepackage{algorithmic}
\usepackage{graphicx}
\usepackage{textcomp}
\usepackage{xcolor}
\def\BibTeX{{\rm B\kern-.05em{\sc i\kern-.025em b}\kern-.08em
    T\kern-.1667em\lower.7ex\hbox{E}\kern-.125emX}}
\begin{document}

\title{Leaf Angle Estimation using Mask R-CNN and LETR Vision Transformer\\
}

\author{\IEEEauthorblockN{Venkat Margapuri}
\IEEEauthorblockA{\textit{Department of Computing Sciences} \\
\textit{Villanova University}\\
Villanova, PA, USA \\
}
\and
\IEEEauthorblockN{Prapti Thapaliya}
\IEEEauthorblockA{\textit{Department of Computing Sciences} \\
\textit{Villanova University}\\
Villanova, PA, USA \\
}
\and
\IEEEauthorblockN{Trevor Rife}
\IEEEauthorblockA{\textit{Pee Dee Research and Education Center } \\
\textit{Clemson University}\\
Florence, SC, USA \\
}
}

\maketitle

\begin{abstract}
Modern day studies show a high degree of correlation between high yielding crop varieties and plants with upright leaf angles. It is observed that plants with upright leaf angles intercept more light than those without upright leaf angles, leading to a higher rate of photosynthesis. Plant scientists and breeders benefit from tools that can directly measure plant parameters in the field; i.e., on-site phenotyping. The estimation of leaf angles by manual means in a field setting is tedious and cumbersome. We mitigate the tedium using a combination of the Mask R-CNN instance segmentation neural network, and Line Segment Transformer (LETR), a vision transformer. The proposed Computer Vision (CV) pipeline is applied on two image datasets, Summer\textunderscore2015-Ames\textunderscore ULA and Summer\textunderscore2015-Ames\textunderscore MLA, with a combined total of 1,827 plant images collected in the field using FieldBook, an Android application aimed at on-site phenotyping. The leaf angles estimated by the proposed pipeline on the image datasets are compared to two independent manual measurements using ImageJ, a Java-based image processing program developed at the National Institutes of Health and the Laboratory for Optical and Computational Instrumentation. The results, when compared for similarity using the Cosine Similarity measure, exhibit $\approx$0.98 similarity scores on both independent measurements of Summer\textunderscore2015-Ames\textunderscore ULA and Summer\textunderscore2015-Ames\textunderscore MLA image datasets, demonstrating the feasibility of the proposed pipeline for on-site measurement of leaf angles.
\end{abstract}


\section{Introduction}
The yields and planting densities of maize in the United States have increased concurrently during the past 50 years \cite{Dzievit}. A comparative analysis of U.S. commercial maize hybrids released since the 1960s revealed that by selecting high yielding hybrids under high planting densities, breeders indirectly selected hybrids with upright leaf angles (LAs) \cite{Dzievit}. The discovery was that upright LAs combined with higher planting densities improve light distribution within the canopy. For instance, modern hybrids intercept 14\% more light than older hybrids \cite{Dzievit}. Leaf Angle Distribution (LAD) is a key parameter that describes the structure of horizontally homogeneous vegetation canopies. It is defined as the probability of a leaf element of unit size to have its normal within a specified unit solid angle \cite{Zou}. LAD affects the manner in which incident photosynthetically active radiation is distributed on plant leaves, thus directly affecting plant productivity. The traditional technique to measure LAD involves the use of mechanical inclinometer, a precision instrument that measures the angle of slope of an object with respect to its gravity by creating an artificial horizon. The use of the mechanical inclinometer makes the process of determining LAD laborious and time-consuming. Other techniques such as 3-D digitizing of individual plant elements using specialized instrumentation \cite{Sinoquet} and laser scanning \cite{Hasoi} are available. However, these techniques are resource-demanding. A feasible, and inexpensive alternative is the use of digital imagery to estimate LA since image analysis preserves the state of the canopy while providing enough insight into the morphometry. Image processing tools such as ImageJ \cite{Schneider} and Adobe Photoshop have built-in angle estimation tools that aid in the estimation of LA. The pitfall is that the operation of the tools is manual which means that the estimations may be subject to user bias and users may spend extended amounts of time to estimate LAs, in case of large datasets. In order to overcome the pitfalls of the manual setup, this paper proposes an automated technique for LA estimation using Mask R-CNN \cite{He} instance segmentation neural network, and Line Segment Transformer (LETR) \cite{Xu}, a vision transformer. 

\section{Related Work}
Dzievit \cite{Dzievit} conducted genetic mapping and a meta-analysis to dissect genetic factors controlling LA variation on maize. Genetic mapping populations were developed using inbred lines B73 (Iowa Still Stalk Synthetic), PHW30 (lodent, expired plant variety protection inbred), and Mo17 (Non-Stiff Stalk) that have distinct LA architectures. The leaf angles were estimated using the ImageJ image processing tool. \newline
\null \quad Herbert \cite{Herbert} described a technique for the estimation of leaf angles using stereo-photogrammetry. The technique permitted accurate measurement of leaf angle and position from several meters away and had sufficient resolution to permit the analysis of complex phenomena such as the effect of leaf shape upon interception of light and photosynthesis. \newline
\null \quad Sinoquet et al. \cite{Sinoquet} proposed a method to measure light interception by vegetation canopies using a 3-D digitizer and image processing software. The 3-D digitizer allowed for simultaneous acquisition of the spatial coordinates of leaf locations and orientations. The software for image synthesis also had the ability to make virtual photographs of the real canopy. The information on light interception was derived from virtual images by using the simple features of image analysis software. \newline
\null \quad Hasoi et al. \cite{Hasoi} estimated the LAD of wheat canopy at different growth stages such as tillering, stem elongation, flowering, and ripening stages by using a high-resolution portable scanning lidar. The canopy was scanned three-dimensionally by optimally inclined laser beams emitted from several measuring points surrounding the canopy and 3-D point cloud images in each stage were obtained. After co-registration of lidar images between different measurement positions, leaves were extracted from the images and each leaf was divided into small pieces along the leaf-length direction. Each of the pieces was approximated as a plane, to which normal (perpendicular) was estimated. The distribution of the leaf inclination angles was derived from the angles of the normal with respect to the zenith. \newline
\null \quad Ryu et al. \cite{Ryu} examined the feasibility of seven techniques such as litterfall, allometry, LAI-2000, TRAC, digital hemispheric photography, digital cover photography, and traversing radiometer system to determine leaf area index across a 9-ha domain in an oak-savanna ecosystem in California, USA. It was shown that the combination of digital cover photography and LAI-2000 could provide spatially representative leaf area index, gap fraction and element clumping index.

\section{Materials and Methods}
\begin{figure}[h]
    \centering
    \includegraphics[width= 3.2in, height=3in, clip]{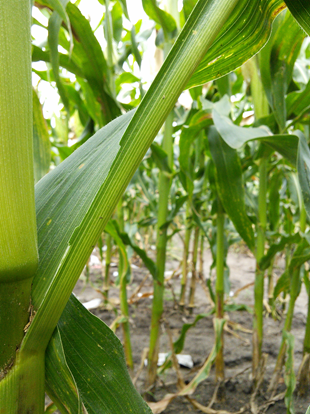}
    \caption{Leaf-Stem Image captured in the Field}
    \label{fig:LeafStemImage}
\end{figure}

Researchers at Iowa State University captured the images of several plants in a field using the Fieldbook android application \cite{Rife}. As for the current work, two datasets namely, Summer\textunderscore 2015-Ames\textunderscore ULA and Summer\textunderscore2015-Ames\textunderscore MLA wherein the former consists of 872 images and latter, 955 images, are used as the experimental datasets. Each of the images in either dataset shows the intersection of the leaf and stem. A sample image is as shown in Fig.~\ref{fig:LeafStemImage}.\newline

\noindent The proposed leaf angle estimation procedure contains two broad steps: \newline
1.	Extraction of region of interest (ROI) \newline
2.	Estimation of Leaf Angle

\subsection{Extraction of Region of Interest (ROI)}
\begin{figure}[t]
    \centering
    \includegraphics[width= 3.2in, height=3in, clip]{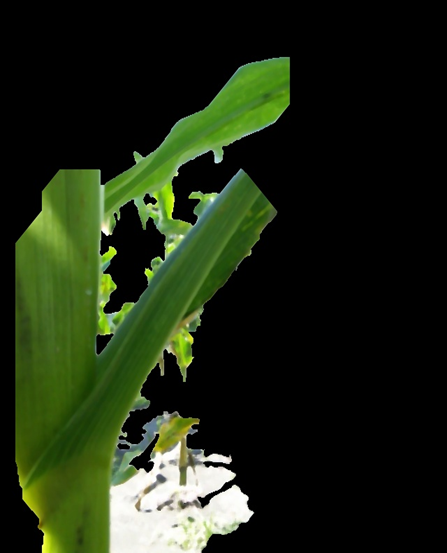}
    \caption{Foreground Extraction using Grabcut Algorithm}
    \label{fig:BotchedExtraction}
\end{figure}

The first step in the extraction of ROI is the detection of leaf and stem within the image. The images captured in the field contain numerous plants that look similar. Hence, it is important to determine the plant of interest and perform a curated extraction of it for the estimation of LA. Conventionally, a feasible technique for foreground extraction is OpenCV’s GrabCut \cite{Talbot} algorithm. However, the results obtained from the application of the GrabCut algorithm on each of the datasets are not satisfactory. While the algorithm captures the foreground, it also captures part of the plants in the background. The captured plants in the background are essentially noise and detrimental to LA estimation. Fig.~\ref{fig:BotchedExtraction} shows the botched foreground extraction performed on a plant image from the Summer\textunderscore2015-Ames\textunderscore MLA dataset.

In order to better perform foreground extraction, the CNN model of Mask R-CNN \cite{He} \cite{Bharati} is employed to extract foreground and identify the region of interest appropriately. Mask R-CNN is a deep neural network aimed to solve the problem of instance segmentation in computer vision. Instance Segmentation \cite{Tian} is the task of precisely identifying the pixels of each of the objects in the image. It is perhaps the hardest of the computer vision (CV) tasks of classification \cite{Lu}, semantic segmentation \cite{Lateef}, object detection \cite{Liu}, and instance segmentation.

\begin{enumerate}
    \item 175 images belonging to each of the Summer\textunderscore2015-Ames\textunderscore MLA and Summer \textunderscore2015-Ames\textunderscore ULA datasets are manually annotated using makesense.ai tool. The tool is free to use and generates annotations in the COCO format that Mask R-CNN is able to consume. A polygonal annotation is drawn where the stem and leaf meet each other. Fig.~\ref{fig:ImageAnnotation} shows a polygonal annotation drawn using makesense.ai on an image with leaf and stem.

\begin{figure}[tbp]
    \centering
    \includegraphics[width= 3.2in, height=4in, clip]{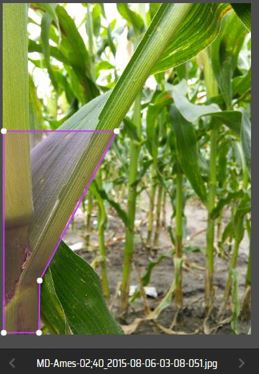}
    \caption{Polygonal Annotation plotted using MakeSense.ai}
    \label{fig:ImageAnnotation}
\end{figure}    

\item The annotated images are trained on the Mask R-CNN network for a total of 20 epochs. The loss of the network after 20 annotations is 0.034. However, the loss is not significant since Mask R-CNN is used to extract the portion of the image where the leaf and stem meet, but not to extract masks that are precise in shape and size. All the annotated images are used for training only. The dataset is not split into training, validation, and test datasets since the goal is not to evaluate the performance of Mask R-CNN, but to extract regions of interest in the image that are later processed to estimate the angle between leaf and stem. Upon training, the trained weights are used to extract the masks of each of the images in the Summer\textunderscore2015-Ames\textunderscore MLA and Summer\textunderscore2015-Ames\textunderscore ULA datasets. Fig.~\ref{fig:LeafStemMask} shows the mask extracted by Mask R-CNN for the plant image shown in Fig.~\ref{fig:ImageAnnotation}.

\begin{figure}[h!]
    \centering
    \includegraphics[width= 3.2in, height=3in, clip]{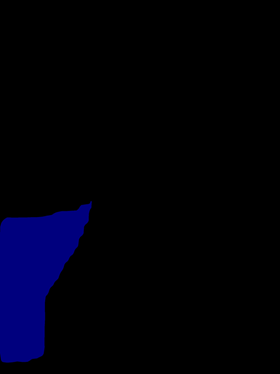}
    \caption{Mask Predicted by Mask R-CNN}
    \label{fig:LeafStemMask}
\end{figure}

\item The predicted mask indicates the region of interest that corresponds to the leaf and stem. In order to extract the region from the original image, the 'bitwise AND' operation is applied on the mask predicted by Mask R-CNN and original image. The resulting image gives the region of interest in the original image. Fig.~\ref{fig:ZoomedImage} shows the region of interest extracted from the original image. Please note that the image is zoomed in for clarity.

\begin{figure}[t]
    \centering
    \includegraphics[width= 3.2in, height=3in, clip]{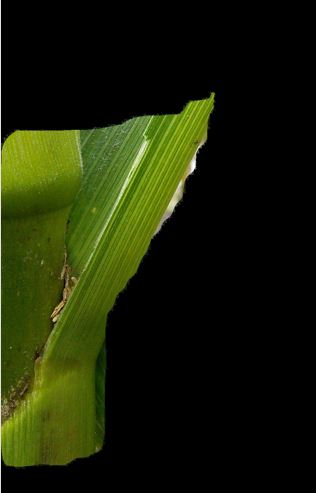}
    \caption{Region of Interest extracted from Original Image}
    \label{fig:ZoomedImage}
\end{figure}
    
\end{enumerate}

\subsection{Estimation of Leaf Angle}
\begin{figure}[t!]
    \centering
    \includegraphics[width= 3.2in, height=3.5in, clip]{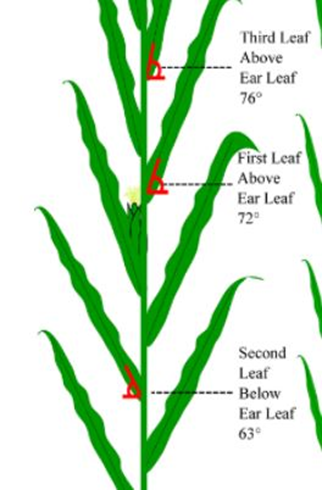}
    \caption{Above Ear and Below Ear Leaf Angles \cite{Dzievit}}
    \label{fig:EarAngles}
\end{figure}

The basis for the estimation of leaf angle is provided by \cite{Dzievit}. The work describes the above-ear and below-ear leaf angles that are measured by plant scientists. Fig.~\ref{fig:EarAngles}, an excerpt from \cite{Dzievit}, shows the exact position at which the angle between leaf and stem is to be measured. While leaves are typically long and tilt in different directions owing to environmental factors, the angle between leaf and stem is measured exactly where the leaf and stem conjoin. \newline
\null \quad The estimation of leaf angle is performed using Line Segment Transformer (LETR), a vision transformer \cite{Xu}. Derived from transformers in the world of Natural Language Processing (NLP), a Vision Transformer (ViT) is a type of neural network that is used to perform CV tasks, such as image classification, and object detection. The basic principle of the transformer used for NLP is the representation of words as sequences. Drawing inspiration from it, ViT works by representing an image as a sequence of patches. A patch is a small rectangular size of the image, typically 16x16 pixels. Each patch that is extracted from the image is represented as a feature vector. The feature vectors are extracted using a convolutional neural network (CNN), such as ResNet-50 or VGG-16. The extracted feature vectors are input to an encoder network. The encoder network is a stack of self-attention \cite{Ramachandran} layers, where self-attention is a mechanism that allows the model to identify and establish long-range dependencies between the patches of images extracted from the original image. The encoder network outputs a sequence of vectors that may be fed into a multi-layer perceptron (MLP) for CV tasks such as classification, and detection. \newline
\null \quad The LETR vision transformer is an adaptation of the Detection Transformer (DETR). The DETR is an object detection model developed by Facebook (Meta) that leverages the transformer architecture. The ResNet architecture is typically preferred as the feature extractor within DETR to extract features from patches of an image. The DETR model follows the encoder-decoder architecture similar to the conventional transformer, with the use of multi head self attention being the key difference between the two. Multi head self attention is the idea where the Attention module comprises N (multiple) attention heads and each attention head repeats its computations multiple times in parallel. All the attention computations are combined to yield a final attention score. Bipartite Matching computed using the Hungarian Algorithm \cite{Sahbani} is used to compute the difference between the ground truth bounding boxes and predicted bounding boxes of each of the objects in the image. The LETR vision transformer follows the DETR model, but is designed specifically for the identification of line segments. The LETR model improves upon the DETR model's ability to detect geometric structures by augmenting the loss term. 

\begin{enumerate}
    \item Xu et al. trained the LETR model on two datasets namely, Wireframe \cite{Huang} and YorkUrban \cite{Denis} using the feature extractors of ResNet-101 and ResNet-50 respectively. The pretrained weights of the LETR model trained on the Wireframe dataset are used for the experiment.
    \item The images from the Summer\textunderscore2015-Ames\textunderscore ULA and Summer\textunderscore2015-Ames\textunderscore MLA datasets are evaluated using the pretrained weights. The line segments output by the model are plotted on evaluation images, as shown in Fig.~\ref{fig:ImageLines}.

    \begin{figure}[t]
    \centering
    \includegraphics[height=3.5in]{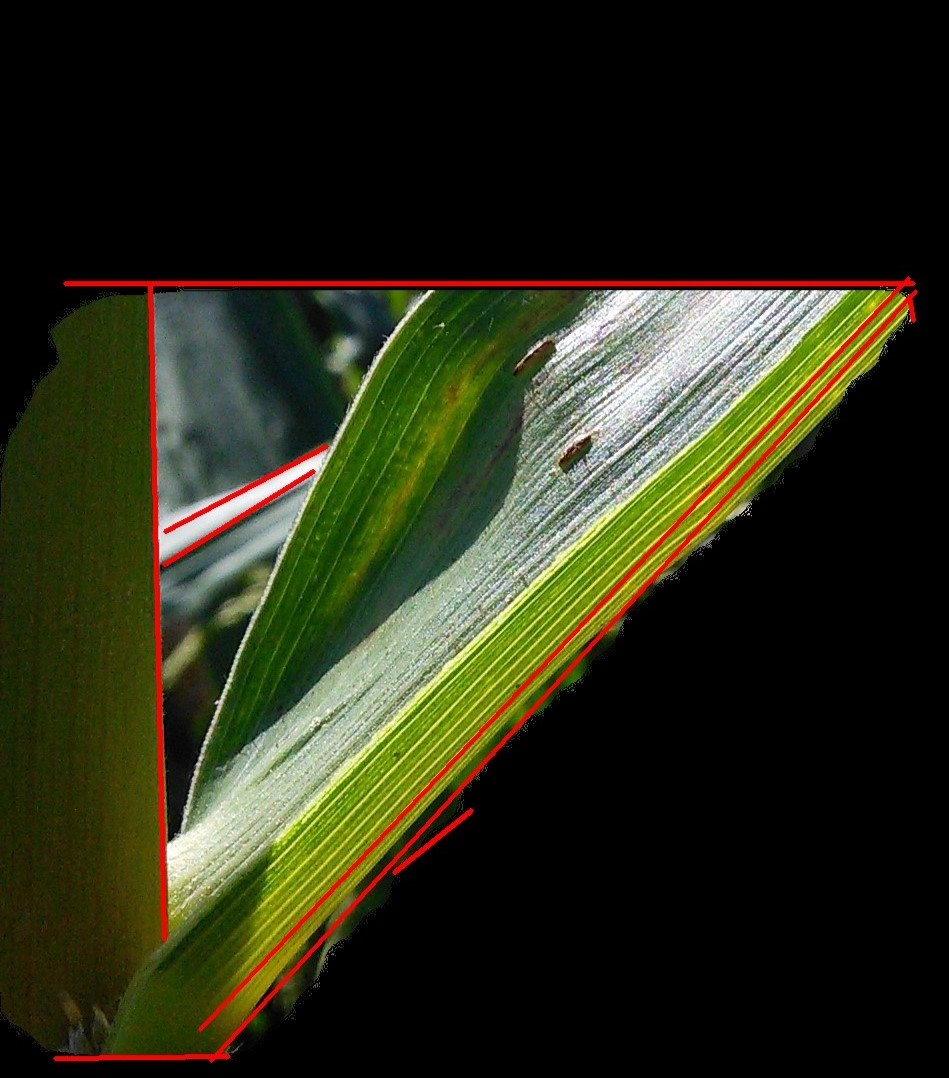}
    \caption{Line Segments identified by LETR (image zoomed in for clarity)}
    \label{fig:ImageLines}
\end{figure}
    \item However, it is the case that several of the detected line segments indicate the stem rather than the leaf, as shown in Fig.~\ref{fig:ImageLines}. The lines on the stem are irrelevant and must be ignored for the estimation of leaf angle. Two constraints, one on the slope and another on proximity to the image boundary are placed on the detected line segments to filter out the ones that don’t belong to the leaf.
    \begin{enumerate}
        \item \textbf{Slope Constraint:} The leaf is generally oriented in comparison to the stem that is almost perpendicular to the ground, in the majority of cases. As a result, line segments whose orientations are between 80 and 90 degrees are ignored, since they are likely to belong to the stem but not the leaf.
        \item \textbf{Boundary Proximity Constraint:} It is empirically determined that the line segments which belong to the leaf are at least 100 pixels apart from the image boundary. The line segments that are closest to the image boundary are those of the stem. Hence, the constraint to ignore line segments that are within 100 pixels of the image boundary works to ignore line segments that belong to the stem. \newline
        Both constraints are required to be satisfied for a line to be ignored. In certain images, the orientation of the leaf is between 80 and 90 degrees. In such cases, without the boundary constraint, the lines pertinent to the leaf are ignored. The line segments that pass the filter are assumed to be present on the leaf.
    \end{enumerate}
    \item Compute the slope of each of the line segments and find the mode of the slopes. Mode refers to the most frequently occurring item among a list of items.
    \item Retrieve the line segments with the most frequently occurring slope for further processing. Only the line segments with the orientation of the mode are considered to eliminate any stray lines that may be detected on the leaf. In the absence of mode i.e. no two line segments have the same slope, the line segment whose slope is the median of all slopes is considered.
    \item Compute the orientation of the line segment made with the horizontal as the tan inverse of slope i.e., angle = $\tan^{-1}$(slope). The orientation made by the line segment is deemed the orientation of the leaf with the stem.
    \end{enumerate}

\section{Impacting Factors}

\subsection{Impact of Image Sharpness}
Image Sharpness describes the clarity of detail in a photo. The aspects of resolution and acutance primarily impact the sharpness of an image. It is imperative that the images be sharp for the line segments to be identified precisely by LETR. Multiple failures are observed with the detection of line segments in images of low sharpness. However, when the sharpness is increased, the line segments are well detected.

\subsection{Multiple Leaf and Stem Detections}
\begin{figure}[t]
    \centering
    \includegraphics[width= 3.2in, height=3.5in, clip]{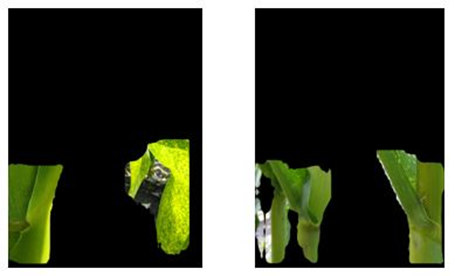}
    \caption{Multiple Leaf-Stem Instances in a Single Image}
    \label{fig:MultipleDetections}    
\end{figure}

In a few cases, the trained Mask R-CNN model detects multiple instances on a given image, some not pertinent to the leaf and stem. Fig.~\ref{fig:MultipleDetections} shows two images where multiple instances are detected. In such cases, the detection of the greatest size (contour) is considered the instance of interest, i.e. leaf and stem. However, it may be flawed in cases where the identified contour is larger than the actual leaf and stem instance in the image. 

\section{Results}

The algorithm is applied to each of the images from the Summer\textunderscore2015-Ames\textunderscore ULA and Summer\textunderscore2015-Ames\textunderscore MLA datasets, and the orientation between leaf and stem is estimated. The results are compared with the leaf angle measurements made by two graduate students at Iowa State University. The students worked independently of each other as they arrived at their estimates using the ImageJ tool. The measurements by the students and the proposed algorithm are compared using a metric known as Cosine Similarity \cite{Vijaymeena}. Cosine Similarity is the cosine of the angle between two vectors that are typically non-zero and within an inner product space.  It is useful to compare the similarity between two vectors represented in a higher-dimensional vector space. Mathematically, it is defined as the division between the dot product of vectors and the product of the magnitude of each vector and is expressed as, similarity = A $\cdot$ B/$\left|A\right|\left|B\right|$ where A and B are the two vectors compared for similarity. The measure is expressed as a value between 0 and 1. \newline
\null \quad On the Summer\textunderscore2015-Ames\textunderscore MLA dataset, the cosine similarity between the proposed algorithm and student 1 is 0.98 which indicates a ten-degree difference in orientation between vectors, and the proposed algorithm and student 2 is 0.99 which indicates an eight-degree difference in orientation between the vectors. On the Summer\textunderscore2015-Ames\textunderscore ULA dataset, the cosine similarity between the proposed algorithm and student 1 is 0.998 which indicates a four-degree difference in orientation between vectors, and the proposed algorithm and student 2 is 0.992 which indicates a seven-degree difference in orientation between the vectors. \newline 
\null \quad Furthermore, an intuitive outlier analysis is performed on the results to gain insight into the similarity of estimation between the different techniques. The rule used to identify an outlier is, “the difference in estimates between the proposed algorithm and a student is greater than eight degrees”. The analysis assumes that the measurements made by the students are accurate. However, the estimations by the students also have a variance between them. The variance in student estimates, given by $\Sigma$(e1-e2)/n where e1 and e2 are estimates made by students 1 and 2 respectively, and n is the total size of the data sample, are 2.57 and 2.07 on Summer\textunderscore2015-Ames\textunderscore MLA dataset and Summer\textunderscore2015-Ames\textunderscore ULA dataset respectively. Hence, the eight-degree range to identify outliers is reasonable. Applying the outlier rule, a total of 31 outliers on the Summer\textunderscore2015-Ames\textunderscore MLA dataset and 16 outliers on the Summer\textunderscore2015-Ames\textunderscore ULA dataset are detected. Among the estimates that are not outliers, a variance of 3.73 degrees and 3.11 degrees is identified between the proposed algorithm and student 1, and the proposed algorithm and student 2 respectively on the Summer\textunderscore2015-Ames\textunderscore MLA dataset whereas a variance of 2.67 degrees and 2.72 degrees is identified between the proposed algorithm and student 1, and proposed algorithm and student 2 respectively on the Summer\textunderscore2015\textunderscore Ames\textunderscore ULA dataset. The results show that the proposed algorithm outputs results that highly co-relate with the manual measurements, demonstrating the feasibility of the technique for leaf angle estimation.

\section{Future Work and Conclusion}
Moving forward, the proposed algorithmic pipeline will be implemented as part of a real-time Android application so plant scientists and individual enthusiasts may leverage the application on a day-to-day basis. The bottleneck for the mobile application is the use of neural network models. The current Mask R-CNN network will have to be adapted so it works on mobile devices using networks such as Mobilenet SSD v1/v2 \cite{Chiu} as the backbone in place of the current ResNet-101 model. The annotation of images is another bottleneck that is laborious and time-consuming. The use of a screen to act as the common background over the plants as images are captured is an efficient way to help identify the location of the plant in the image. \newline
\null \quad Overall, the proposed technique demonstrates the idea of estimating the angle between leaf and stem using instance segmentation neural networks and vision transformers, a task that is laborious and cumbersome when performed manually. We will implement the leaf angle estimation as a feature in the FieldBook \cite{Rife} Android application. The estimation of leaf angles using automation is novel and has the potential to turn into a one-of-a-kind utility for the agricultural community.

\end{document}